\documentclass[lettersize,journal]{IEEEtran}

\usepackage{amsmath,amsfonts}
\usepackage{algorithmic}
\usepackage{algorithm}
\usepackage{array}
\usepackage[caption=false,font=normalsize,labelfont=sf,textfont=sf]{subfig}
\usepackage{textcomp}
\usepackage{url}
\usepackage{verbatim}
\usepackage{graphicx}
\usepackage{makecell}
\usepackage[bottom=76pt, top=57pt, left=45pt, right=45 pt ]{geometry}
\usepackage{comment}
\usepackage{xcolor}
\usepackage{color}
\usepackage{hyperref}
\usepackage{wrapfig,lipsum,booktabs}
\usepackage{cite}
\newcommand{\hakan}[1]{\textcolor{black}{#1}}
\newcommand{\erhan}[1]{\textcolor{black}{#1}}

\begin{document}

\title{Cross-Embodied Affordance Transfer through Learning Affordance Equivalences}

\author{Hakan Aktas$^{1,2}$, Yukie Nagai$^{2}$, Minoru Asada$^{3,4}$, Matteo Saveriano$^{5}$, Erhan Oztop$^{3,6}$, Emre Ugur$^{1}$
\thanks{$^{1}$Hakan Aktas and Emre Ugur are with Computer Engineering Department, Bogazici University
        {\tt hakanwhitestone@gmail.com}
$^{2}$Hakan Aktas and Yukie Nagai are with IRCN, the University of Tokyo
$^{3}$Minoru Asada and Erhan Oztop are with SISREC, Osaka University
$^{4}$Minoru Asada is also affiliated with the International Professional University of Technology in Osaka 
$^{5}$ Matteo Saveriano is affiliated with University of Trento,
$^{6}$Erhan Oztop is also affiliated with Computer Engineering Department, Ozyegin University.
}}



\maketitle

\begin{abstract}

Affordances represent the inherent effect and action possibilities that objects offer to the agents within a given context. From a theoretical viewpoint, affordances bridge the gap between effect and action, providing a functional understanding of the connections between the actions of an agent and its environment in terms of the effects it can cause. In this study, we propose a deep neural network model that unifies objects, actions, and effects into a single latent vector in a common latent space that we call the affordance space. Using the affordance space, our system can generate effect trajectories when action and object are given and can generate action trajectories when effect trajectories and objects are given. Our model does not learn the behavior of individual objects acted upon by a single agent. Still, rather, it forms a `shared affordance representation' spanning multiple agents and objects, which we call Affordance Equivalence. Affordance Equivalence facilitates not only action generalization over objects but also Cross Embodiment transfer linking actions of different robots. In addition to the simulation experiments that demonstrate the proposed model's range of capabilities, we also showcase that our model can be used for direct imitation in real-world settings. 

\end{abstract}


\section{Introduction}
Ecological psychologist James Gibson coined the term affordances in his theory of visual perception to refer to the directly perceivable action possibilities provided by the environment to embodied agents \cite{gibson1986}. Affordances have been an influential concept in robotics as they encode the tight coupling between the agent and the environment. As such, robotic systems, which may have robot-specific perceptual and motor capabilities, would definitely benefit from affordances to efficiently perceive the available actions and act accordingly. Affordances \cite{csahin2007afford} has been extensively studied in robotics research in the last two decades.  Initially used as a source of inspiration from ecological \cite{gibson1986} and developmental \cite{gibson1969principles} psychology, where paradigms such as direct perception, emergence, and animal-environment coupling have initially been applied in robotic research \cite{uugur2010traversability,kroemer2012kernel}, later the affordance concept has been stretched to agent-free perception \cite{myers2015affordance}, planning \cite{aktas2024multi} or high-level cognition \cite{ugur2011goal,awaad2013affordance}. However, as noted in \cite{Zech2017}, ``solving the correspondence problem'', that is, mapping affordances between different agents, remains an important challenge in robotics affordance research.

A large body of affordance literature has focused on the challenge of self-supervised learning \cite{Zech2017} of affordances through a process shaped by goal-free exploration and interaction with the environment, where the agent observes and processes the effects of its own actions. In these studies, affordances refer to the relations among objects, actions, and effects and have generally been represented by models that predict effects given the objects and actions\cite{Jamone2016,taniguchi2023world}. 
Here, we adopt and extend the conceptual affordance formalization of \cite{csahin2007afford} and develop a deep learning architecture conforming to the formalization to form latent representations via learning and establish affordance equivalences that equip agents with non-trivial capabilities such as effect-based imitation and cross-embodied skill transfer.

\erhan{In our earlier studies, we have showed the effectiveness of latent representation blending for multimodal learning \cite{seker2022imitation} and for establishing correspondences between morphologically different robots\cite{aktas2024correspondence}. However, these studies did not address the method of systematically incorporating the interacted object information into the blending framework. In this study, we take the blending concept one step further by incorporating static elements, such as object depth maps, alongside time series information to enable the model to predict time series data from static inputs as well as to enable it to generate consistent static output based on input time series data. This capability has allowed us to computationally represent affordance relations that include both static and time series components, in which  action and effect components are represented by time series data, whereas the interacted object information is captured by static depth images.}

Overall, the developed architecture forms a framework overarching multiple agents and their high-dimensional temporal sensorimotor experience, enabling inference with missing affordance components in high-dimensional affordance spaces. For example, given a robot and its action, our system can infer the object properties (depth image) in order to obtain the desired effect. Alternatively, given a desired effect on an object, our system can infer the required robot and action.

\erhan{To the best of our knowledge, the proposed system is the first to achieve the complete formation of affordance equivalences with all affordance components, including robotic embodiments. This also showcases the benefits of representing these equivalences in a unified affordance format through complex robotic experiments. For instance, the proposed framework effectively addresses the complex issue of transferring affordances between robots with varying embodiments and capabilities. Key attributes of our system include}

\begin{itemize}
    \item It can form a common latent affordance representation space from high-dimensional robot-action-effect-object interaction data following affordance equivalences.
    \item  It can be used for cross-embodied learning and transfer of affordances between different agents
    \item It enables direct imitation in the real world using affordance equivalences.
\end{itemize}

\erhan{To elucidate the merits of our model, we have assessed its ability to learn various affordances and their corresponding equivalences through multiple experiments, each focusing on different aspects of our model, which we report in Experimental Results. In brief, we explore the insertability-related object equivalences to demonstrate our model's adaptability to different input modalities, including force,  and show its superiority to a standard baseline (Subsection~\ref{section:insetion}). We also investigate  the latent space structure as a function of learning time. Then, the graspability affordance and its cross-embodied transfer, highlighting the system’s behavior in a multi-agent context, are considered where  the actions of agents are not equivalent for all the objects since objects offer different affordances for different agents' actions (Subsection~\ref{section:grasp}). Further,  our model's cross-embodied affordance transfer ability with novel objects is shown in the subsection \ref{section:push}. Lastly, a real-world experiment showcasing how the \emph{effect} channel of the model can be utilized for direct imitation is presented (Subsection~\ref{section:real-robot}.}

\section{Related Work}

\label{sec:relatedwork}

\textbf{Affordances} refer to the action possibilities offered to agents\cite{gibson1986}. Since the conception of the term, it has been widely influential in various fields, including ecological and developmental psychology, human-computer interaction, design, artificial intelligence, and robotics\cite{jamone2016affordances}. Yet, it has significantly deviated from Gibson's original intention in both psychology  \cite{chong2020evolution} and robotics \cite{csahin2007afford, min2016affordance, yang2023recent}. One important direction in affordance research loosens the coupling between the agent and affordances by either annotating the affordances manually \cite{myers2015affordance} or directly learning affordances from human performances \cite{saponaro2019beyond,bahl2023affordances}. In order to tighten this coupling, others ground the existing affordances in the robots' sensorimotor experience in a separate process \cite{ahn2022can}. In this paper, we study self-supervised learning of affordances through a robot's interaction with the environment. Previous self-supervised studies focused on learning and predicting the effects of robot actions given object and action information \cite{ugur2014emergent,ugur2016emergent,ugur2015refining,seker2019deep}. Few affordance studies deal with predicting actions and action parameters in a goal-oriented manner \cite{ugur2011going}. Going beyond effect prediction, in this work, we let the robot collect information related to objects, actions, and the generated effects to make predictions in any channel (object, effect, or action), given one or more of them similar to \cite{montesano2008learning}. While \cite{montesano2008learning} can only model low-dimensional object/action/effect spaces, our system uses the representational robustness of Conditional Neural Processes \cite{garnelo2018conditional} to process high-dimensional temporal data as well. In addition to action information, our data includes interactions with different robots, and our system is able to handle different robots as inputs and outputs.

\textbf{Motion and sensorimotor encoding} has been a topic studied by many in recent years \cite{asfour2008imitation,amor2012generalization,paraschos2018using}. Several different encoding methods have been proposed in various studies. Some were based on statistical modeling and dynamic systems \cite{calinon2016tutorial,huang2019kernelized,zhou2017task}, while others used Locally Weighted Regression to learn the environment parameters \cite{kramberger2017generalization,ude2010task}. Locally Weighted Projection Regression \cite{vijayakumar2000locally}, Gaussian Mixture Models \cite{calinon2009learning,pervez2018learning} and Hidden Markov Models \cite{chu2013using,girgin2018associative,ugur2020compliant} are also utilized to learn motion patterns from demonstrated actions.  In recent years, neural networks have also been widely used to learn motion primitives using complex data \cite{gams2018deep,pervez2017learning,xie2020deep}. In a previous study, the Conditional Neural Movement Primitives \cite{seker2019conditional} was introduced as a deep LfD architecture that can be used to learn complex motion trajectories. It can learn low-dimensional relations from a few samples and high-dimensional relations from larger datasets. This stochastic method is based on Conditional Neural Processes \cite{garnelo2018conditional}, which has the ability to construct full trajectories using any given points on arbitrary time steps. 
Because of these capabilities, we used the CNMPs as a backbone system for modeling the temporal sensorimotor data in our model.

\begin{figure*}[t]
\centerline{\includegraphics[width=0.75\linewidth]{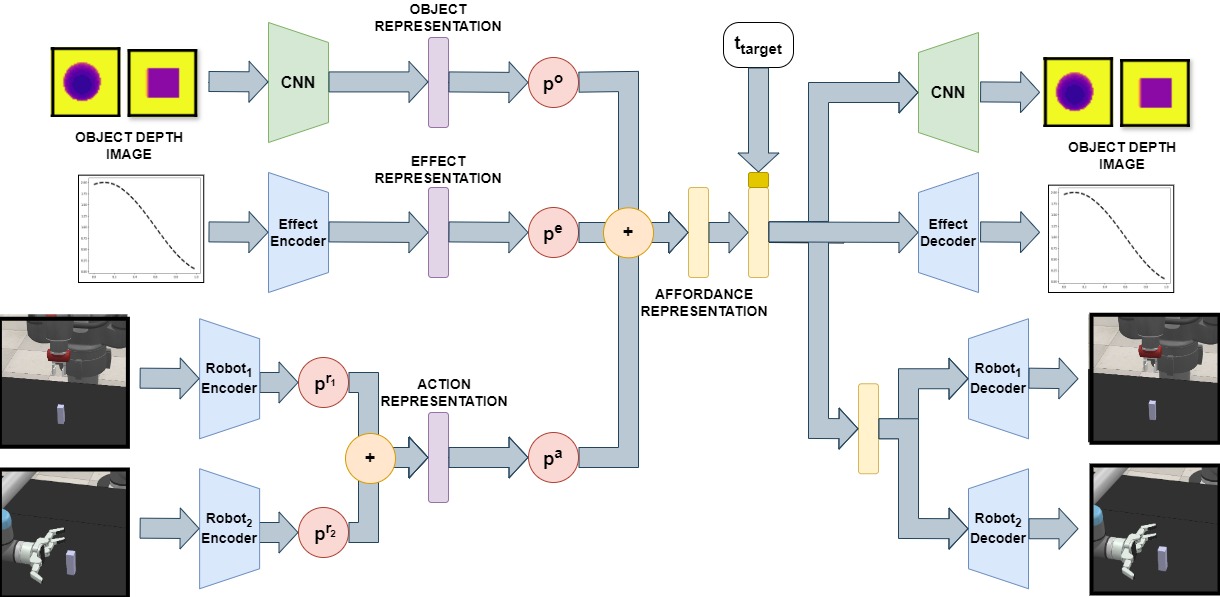}}
\caption{Overview of the proposed model (refer to the text for the details of training and generation). The model has a channel for each element of the affordance tuple. The action channel can be comprised of several channels for different agents involved. The system is able to predict missing elements of the affordance tuple given information about the other components.}
\label{figure:method}
\end{figure*}


\textbf{Cross Embodiment Transfer:} The previous works on cross-embodiment transfer aim to transfer or generalize collected robot motion trajectories to various robots with different embodiments \cite{taylor2009transfer,zhu2023transfer}. These studies include transfer in and in between domains such as transfer to robots with different morphologies in simulation\cite{yu2017preparing,chen2018hardware}, transfer to different robots in real world\cite{yang2023polybot}, from robots in simulation to real robots\cite{christiano2016transfer,zhang2020learning}. However, these studies do not address the affordances of the objects. Several methods have been introduced to use cross-embodiment transfer to expedite the training by using information from other agents. When the policy of the source agent is known, several studies showed that reinforcement learning (RL) can be used to fine-tune the value function\cite{konidaris2006autonomous}, the visual encoder \cite{rusu2017sim}, or the policy \cite{liu2023meta} of the target robot. In a previous study, we also showed that cross-embodiment transfer of learned representations can be used to bootstrap RL \cite{akbulut2021acnmp}. \cite{aktas2024correspondence} also proposed a method that can align and transfer the actions of the robots if a common task space can be established between them. In this study, we further address the transfer of affordances of objects with different shapes between robots with different physical and motor capabilities.
\section{Method}

\label{sec:method}
\subsection{Affordance Formalization}

In this paper, we use an affordance representation inspired by~\cite{csahin2007afford}. The main differences are agent is added as a distinct component besides actions, and we exploit an object-centric approach to define the affordances, while the previous work uses object-free entities and behaviors. In \cite{csahin2007afford}, entities are used to represent any being interactable by the agents including environments or other agents. We restrict this by only focusing on objects. Furthermore, the term behavior represents more general interactions of the agents.
 Instead, we use the term action to represent only the interaction with the object. We formalize and represent the affordances as
\begin{equation}\label{eq:aff-notation}
    (\tt{effect},(\tt{object},\tt{action})),
\end{equation}
where we define the $\tt{object}$ as the entity the robot/agent interacts with, and the $\tt{effect}$ as the resulting change caused by the $\tt{action}$ when it is executed on the given object. Action can be any motion the robot/agent executes. Using this representation, we also define several equivalences. When the same action is applied to different objects and the same effect is observed, we state that their affordance for that action and effect are equivalent. We formalize this equivalence as 
\begin{equation}\label{eq:aff-object-eq}
  (\tt{effect},(\biggl\{   \begin{array}{l}
            \tt{object}_1\\
            \tt{object}_2
        \end{array}     \biggl\}, action).
\end{equation}
We also define the affordance of two actions as equivalent if the effects caused by that action on the given object are the same. We formalize action equivalence as 
\begin{equation}\label{eq:aff-action-eq}
  (\tt{effect},(\tt{object},\biggl\{   \begin{array}{l}
            \tt{action}_1\\
            \tt{action}_2
        \end{array}     \biggl\} ).
\end{equation}
When multiple agents/robots are involved, the notation for the affordance representation becomes
\begin{equation}\label{eq:aff-notation-agent}
    (\tt{effect},(\tt{agent},(\tt{object},\tt{action}))) .
\end{equation}
In this case, the actions of two different agents can also be equivalent and can be formalized differently based on the actions. If the actions executed by the agents are the same, then we have
\begin{equation}\label{eq:aff-agent-eq}
  (\tt{effect},( \biggl\{   \begin{array}{l}
            \tt{agent}_1\\
            \tt{agent}_2
        \end{array}     \biggl\},(\tt{object}, \tt{action})).
\end{equation}
An example of this case would be
\begin{equation}\label{eq:aff-agent-eq-ex}
  (\tt{lifted},( \biggl\{   \begin{array}{l}
            \tt{robot}_1\\
            \tt{robot}_2
        \end{array}     \biggl\},(\tt{cuboid},\tt{lift} )),
\end{equation}
where we assume that both $\tt{robot}_1$ and $\tt{robot}_2$ can lift the given object ($\tt{cuboid}$) with their $\tt{lift}$ action. On the other hand, if the actions performed by the agents are different, the formalization becomes,
\begin{equation}\label{eq:aff-agent-action-eq}
  (\tt{effect}, \biggl\{   \begin{array}{l}
            \tt{agent}_1\\
            \tt{agent}_2
        \end{array}    , \biggl\{\tt{object}, \begin{array}{l}
             \tt{action}_1  \\
             \tt{action}_2 
        \end{array}  \biggl\}  \biggl\} ).
\end{equation}
An example of this case would be
\begin{equation}\label{eq:aff-agent-action-eq-ex}
  (\tt{moved}, \biggl\{   \begin{array}{l}
            \tt{robot}_1\\
            \tt{robot}_2
        \end{array}    , \biggl\{ \tt{cuboid}, \begin{array}{l}
             \tt{push}  \\
             \tt{pull} 
        \end{array}  \biggl\} \biggl\}  ),
\end{equation}
where we assume that $\tt{robot}_1$ can move the $\tt{cuboid}$ object by using a $\tt{push}$ action and $\tt{robot}_2$ can move it by using a $\tt{pull}$ action.

\subsection{Proposed Model}

\begin{figure*}[t]
\centerline{\includegraphics[width=1\linewidth]{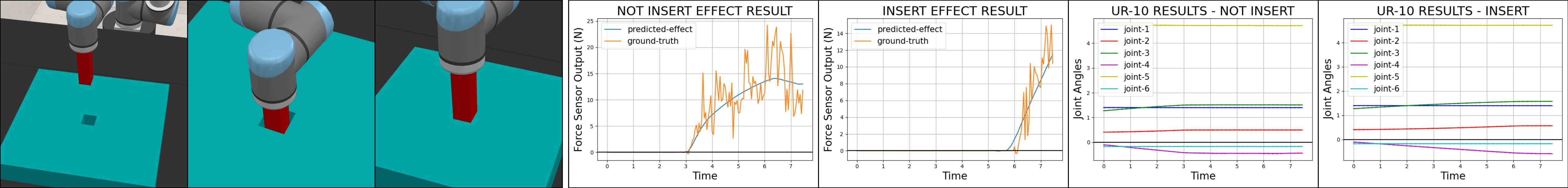}}
\caption{
The setup of the experiment conducted in Section \ref{section:insetion} (on the left) and exemplary results of the same experiment. As seen from the effect plots in the middle, when the opening is insertable the force readings peak earlier (middle left) than when it is not insertable (middle right). The plots on the right show the action generation results where the solid lines show the generated trajectories and the dashed lines show the ground truth values (dashed lines are hard to see since they overlap). 
}
\label{figure:insertion-setup}
\end{figure*}

This study proposes a neural network model that learns affordances by observing the {\it effects} of {\it robots}'s {\it actions} on {\it objects} and is designed to reconstruct {\it object}, {\it effect}, {\it action}, and {\it robot} information when one or more of these is missing.
In an earlier work \cite{aktas2024correspondence}, we showed that a multi-channel CNMP \cite{seker2019conditional}  model can be used to find correspondences between robots with different morphology. In this study, we generalize the method to include object affordances. The interactions of the robots with the objects are used to form an affordance representation space, as shown in Fig. \ref{figure:method}. The depth images of  \erhan{objects are used to relay} 
object information.

To represent the effect, we used trajectories that hold information about the \erhan{movement patterns of the objects.} 
Similarly, for the action, we used trajectories that encode the robot actions such as joint angle trajectories. Following the CNMP \erhan{framework}, the timestamps of the observations are also used for encoders with trajectory inputs. For effect and robot encoders, fully connected layers and for object encoders, convolutional layers followed by fully connected layers are used. The encoders map them to individual object, effect, and action latent vectors of the same size. Next, to provide the capability to reconstruct the missing channel, following \cite{seker2022imitation}, these representations are blended by taking random convex combinations to yield a shared multi-model multi-agent affordance representation. The affordance representation is expected to encode the object-robot-action-effect relations robustly so that even though one or two of these components are missing, the affordance representation should be sufficient to reconstruct the missing components through the corresponding decoders. Formally, let $A$ be the dataset that contains the affordance instance dataset:
\begin{equation}
\label{eq1}
    A_{j} = (\{a_{t}^{R_j},e_{t},o,t\}_{t=0}^{t=1})_{j}
\end{equation}
where $0 \leq t \leq 1$ is normalized time, $e_t$ denotes the observed effect at time $t$, $o$ represents the object image, and $a_t^{R_j}$ represents the action/motor output at time $t$ pertaining to agents $R_j = \{Agent_{1}, Agent_{2},...,Agent_{k}\}$. Index $j$ indicates the sample number with $0 \leq j \leq m$ where $m$ is the size of the dataset. 
Note that different agents have different motor capabilities and therefore environment provides different affordances to those agents. Hence for fulfilling a common task, individual agents may need very different motor outputs, which we map them to a shared action representation through a convex combination scheme as in \cite{aktas2024correspondence}. For this, we first obtain individual latent representations for each agent with

\begin{equation}
 \label{eq:1} L_{i}^{a_{r_j}} = E^{a_{r_j}}((t_i,a_{t_i}^{r_j})|\theta^{a_{r_j}}) \quad r_j \in R_j 
\end{equation}

where $E^{a_{r_j}}$ is an action deep encoder for agent $r_j$ with weights $\theta^{a_{r_j}}$, and $L_i^{a_{r_j}}$ is the action latent representation of agent $r_j$ constructed using the given observation. $i$ is the index of the sampled observation. Then, these sampled representations are averaged for each agent with  

\begin{equation}
L^{a_{r_j}} = \frac{1}{n} \sum_{i}^{n} L_{i}^{a_{r_j}} \quad r_j \in R_j
\end{equation}

where $n$ is the number of sampled observations during this iteration. Then a convex combination is taken to obtain the common latent action vector 
\begin{equation}
L^a = \sum_{r_j}^{R_j} p^{r_j} L^{a_{r_j}}
\end{equation}

where $p^{r_j}$ are real positive values summing to $1$.
Effect and  object latent representations are obtained similarly: 
\begin{equation}
L_{i}^e = E^{e}((t_i,e_{t_i})|\theta^e), \quad L^{e} = \frac{1}{n} \sum_{i}^{n} L_{i}^{e}
\end{equation}
\begin{equation}
 L_{i}^o = E^{o}((t_i,o_{t_i})|\theta^o), \quad L^{o} = \frac{1}{n} \sum_{i}^{n} L_{i}^{o}
 \end{equation}

Here, $E^{e}$ and $E^{o}$  denote the effect and object deep encoders respectively. The obtained latent action, effect, and object representations are then combined using yet another convex combination step:
\begin{equation}
    L^F = p^{a} L^a + p^{e} L^e + p^{o} L^o  \quad 0\le p \le 1, p^{a}+p^{e}+p^{o} = 1
\end{equation}
where $L^F$ represents the affordance tuple that can be used to reconstruct the trajectories used to form it. After the representations are merged into one, this representation is decoded to obtain target distributions on $t_{target}$ for the action of all agents, objects, and effects: 
\begin{equation}
(\mu_{t_{target}}^{a_{r_j}},\sigma_{t_{target}}^{a_{r_j}} ) = Q^{a_{r_j}}((L^F,t_{target})|\phi^{a_{r_j}}) \quad r_j \in R_j
\end{equation}
\begin{equation}
    (\mu_{t_{target}}^e,\sigma_{t_{target}}^e ) = Q^e((L^F,t_{target})|\phi^e)
\end{equation}
\begin{equation}
(\mu_{t_{target}}^o,\sigma_{t_{target}}^o ) = Q^e((L^F,t_{target})|\phi^o)
\end{equation}

where $Q^r$ is a deep decoder with weights $\phi^r$ that constructs distributions with mean $\mu_{t_{target}}^r$ and variance $\sigma_{t_{target}}^r$ for the robot $r$. The learning goal of our system is to obtain  accurate distributions for given condition points (observations given to the system from the inputs of the encoders) and the target point (the concatenated time $t_{target}$ which is the time stamp of the desired observation point as output), and therefore, the loss is defined similarly to \cite{seker2019conditional}:
\begin{equation}
    \mathcal{L}(\theta, \phi) = - \log P(y_{j} | \mu_{j}, \mathrm{softmax}(\sigma_{j})).
 \label{eq:cnmp_loss}
\end{equation}

Our grand goal is to enable a seamless transfer of affordances between agents even though an affordance relation has been demonstrated only to one of the agents. After sufficient training with affordances known to all agents, the demonstration of the affordance known only by one agent can also be used for training. Note that the system should continue to be trained using the old data alongside the new one to prevent catastrophic forgetting. After sufficient training with the new data, the system can produce the other agent's action and the corresponding action's effect when conditioned with the new object if the affordance is shared.

\begin{figure*}[t]
\centerline{\includegraphics[width=0.7\linewidth]{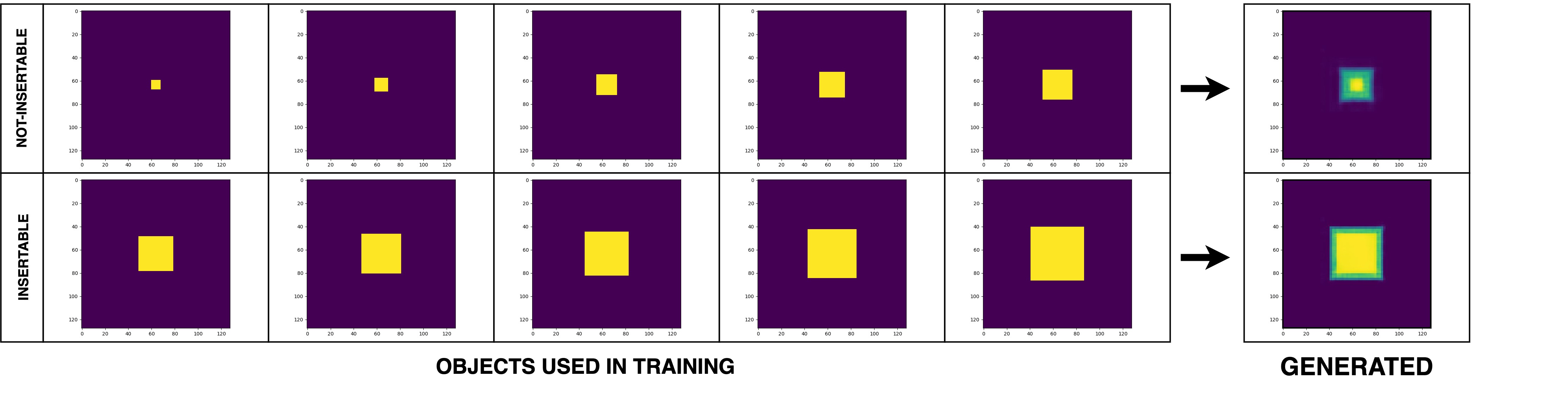}}
\caption{
The depth images of non-insertable (top row) and insertable (bottom row) openings used for training in the experiment in Section \ref{section:insetion} and the generated depth images reconstructed using common affordance representations (on the right).
}
\label{figure:insertion-depth-images}
\end{figure*}

\section{Experimental Results}
\label{sec:experiments}


\subsection{Learning Object Equivalences for Insertability}\label{section:insetion}

\begin{table}[b]
\centering
\caption{ The average RMS differences of the generated trajectories from the ground truth when different input configurations are used. The image error values are pixel errors, action errors are differences in joint angles (radians), and effect errors are in newtons.}
\label{tab:insert-results}
\scalebox{0.8}{
    \begin{tabular}{|c|c|c|c|c|l|l|l|}\hline \hline 
         \makecell{input\\ configuration}&  \makecell{object \\ +effect \\+action}& \makecell{ object \\+effect}&  \makecell{object \\+action}& \makecell{ effect \\ +action} & object& effect&action\\ \hline 
         \makecell{image \\ error \\ (pixels) }&  0.107&  0.105&   0.106&   0.106&  0.1051& 0.106&0.151\\ \hline 
         \makecell{effect \\ error \\ (newtons)}&  2.195&   2.119&  2.086&  2.812&  2.212& 2.754&4.635\\ \hline 
         \makecell{action\\  error \\ (radians)}&  0.00544&  0.0057&  0.00504& 0.00611& 0.00613&  0.006& 0.0206\\ \hline
    \end{tabular}}
\end{table}

This experiment aims to verify that our model can encode a widely used affordance, namely insertability affordance,  and represent the object equivalences in its common latent representation space. A single robot (UR-10) is used in a simulated environment \cite{coppeliaSim}. UR-10 is equipped with a fixed-sized red rod, as shown on the left side of Figure \ref{figure:insertion-setup}. The task is defined as inserting the rod into the opening on the planar surface of a table. The sizes of the openings vary across experiments while their shape is always square.
The action and object are encoded as the robot arm's joint position trajectory and the depth image of the surface centered around the opening, respectively. The force sensor readings, obtained from a sensor placed at the wrist of the robot, are used as effect information to show that our model can use different modalities as effect rather than only positional data. The depth images of 8 different-sized openings are used for training (4 insertable and 4 non-insertable), and 2 are used for testing (1 insertable and 1 non-insertable) to show that our model can generalize to unseen opening sizes. The two images used in the test set have opening sizes selected within the range of insertable and non-insertable openings and not in the decision boundaries. The depth images used in training can be seen on the left of Figure \ref{figure:insertion-depth-images} (all except the ones on the right). 

\begin{figure}[b]
    \centering
    \includegraphics[width=0.7\linewidth]{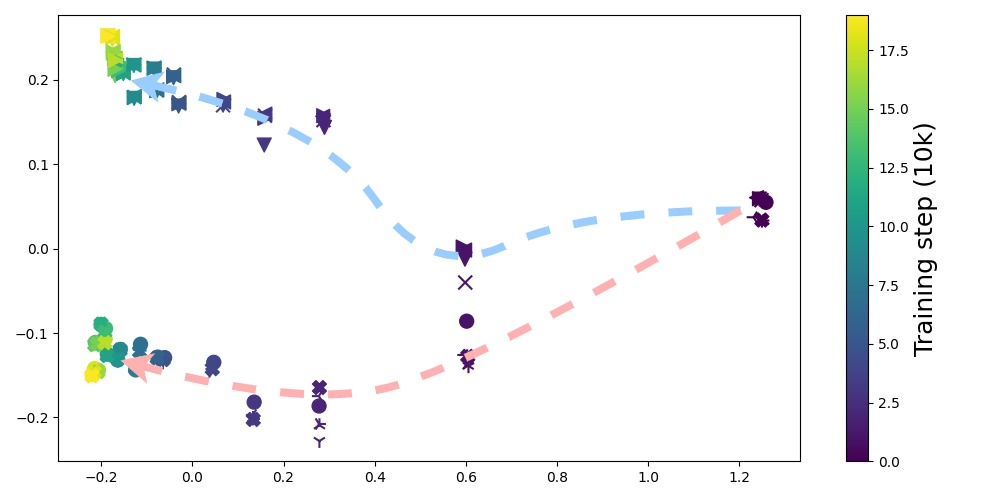}
    \caption{Latent space analysis of the formed affordance representations of the experiment in Section~\ref{section:insetion}. Each different shape shows a different object. It can be seen that as training progresses, half of the objects converge to one point (insertable), and the other half converges to the other point (not insertable).}
    \label{figure:latent-space}
\end{figure}

\begin{table}[t]
\centering
\caption{Generation accuracy comparison of our model and the method used in \cite{zambelli2020multimodal}. It can be seen that our model outperforms the baseline in predicting all affordance components.}

\label{tab:baseline}
\scalebox{1}{ 
\begin{tabular}{|c|c|l|l|l|l|}\hline \hline 
         &  \makecell{Ours}& NLL-w/o t&MSE-w/o t& NLL-w/t&MSE-w/t\\ \hline 
         \makecell{image\\  error \\ (pixels)}&  \textbf{0.11}& 0.12& 0.12 
& 
0.15&0.12
\\ \hline 
         \makecell{effect \\ error \\ (newtons)}&  \textbf{2.68}& 6.30& 6.424 
& 6.09&5.38
\\ \hline 
         \makecell{action \\ error \\ (radians)}&  \textbf{0.007}& 0.084& 0.104 & 
0.034&0.0277\\ \hline
    \end{tabular}}
\end{table}

The predicted and ground truth trajectories for effect and actions are given in Figure \ref{figure:insertion-setup}.
As shown, when the opening is insertable, the force readings start to rise at the end of the execution because the tip of the rod touches the table at the end. When the opening is not insertable, the force readings start to rise in the middle of the execution because the tip of the rod touches the edges of the opening. Due to this difference in motion, the action trajectories are also slightly different, as seen in the figure. We further analyzed the performance of our system by systematically changing the missing input channels. Table \ref{tab:insert-results} provides the reconstruction errors for different input channels. The error values are the mean values of all possible novel input observation combinations. For instance, input configuration $object+effect$ corresponds to the setting where $p^o$ and $p^e$ are set to $0.5$, and $p^a$ is set to 0. The image error values are pixel errors, action errors are differences in joint angles (in radians), and effect errors are in newtons. As shown, the reconstruction errors do not vary significantly when different configurations are used. 
The system can generate the desired output even when only one affordance component is available since the data for insertables and non-insertables are disjoint in this experiment. One can see the differences between the predicted and the ground truth trajectories in Figure \ref{figure:insertion-setup}. 
One can also see that the image reconstruction errors are the same for almost all conditions. 
The depth images generated from the common representations can be seen on the right of Figure \ref{figure:insertion-depth-images}.

Next, we analyzed the evolution of the formed common representation and visualized it  in Figure \ref{figure:latent-space}. The plot shows how the affordance representation changes as training progresses. To reconstruct the affordance representations, we set the weights of the system (action, effect, and object) to 0.33. We sampled several novel points along the effect and action trajectories as input to the effect and action encoders. We used all ten openings shown in the figure by different markers. For visualization purposes, Principal Component Analysis (PCA) is used to decrease the dimension of each affordance representation to two.  The color bar shows the training progress. As shown, at the beginning of the training (on the right side of the plot), the representations are closer together due to initialization, and as the training progresses, they diverge, and two groups converge into two different points. Note that some markers have very similar projections and thus are overlapping. This shows that rather than learning different affordance representations for each $(effect,(agent,(object, action)))$ tuple, our model learns the equivalence between the objects. The two equivalence classes learned by our model in this experiment can be succinctly given with the adopted notation as $(inserted,(UR-10,(<insertable\text{-}openings>, insert))$ and $(not\text{-} inserted,(UR-10, (<non\text{-}insertable\text{-}openings>, insert))$
where $<insertable\text{-}openings>$ represents the openings that are insertable.

Finally, we compared the reconstruction performance of our method with a baseline. While our method uses blending to fuse different latent representations, another approach is concatenating the representations, as shown in \cite{zambelli2020multimodal}. We constructed a similar architecture that can make next-step predictions and compared it to our model. We also used the same method in \cite{zambelli2020multimodal} to drop representations during training. We trained the model with both our system's loss function and Mean Square Error (MSE). Furthermore, since our model uses time steps of states in its inputs, we added time to the inputs of the baseline model and measured its performance. The results are given in Table \ref{tab:baseline} 
\erhan{ where the columns labeled with NLL-wo/t and NLL-w/t, the models are trained with negative log-likelihood loss, while the columns labeled with MSE-wo/t and MSE-w/t show the results with mean squared error loss. Time steps are appended to the state inputs in NLL-w/t and MSE-w/t cases, whereas for NLL-wo/t and MSE-wo/t cases time is not included in the input.}
The table shows that adding time to the input decreased the action errors of the baseline model. However, our model still significantly outperforms others in both effect and action prediction.

\subsection{Learning Agent Equivalences for Graspability }\label{section:grasp}

The previous experiment has shown that our model can encode single-agent affordances and reconstruct the missing affordance components. This experiment aims to evaluate the capabilities of our model in encoding multi-agent affordances and cross-embodied affordance learning. Additionally, while only a single action was included in the previous \erhan{experiment, here we consider different actions from the robots involved to assess the ability of our system to capture agent-action interaction.}


\begin{figure}[t]
    \centering
    \includegraphics[width=0.7\linewidth]{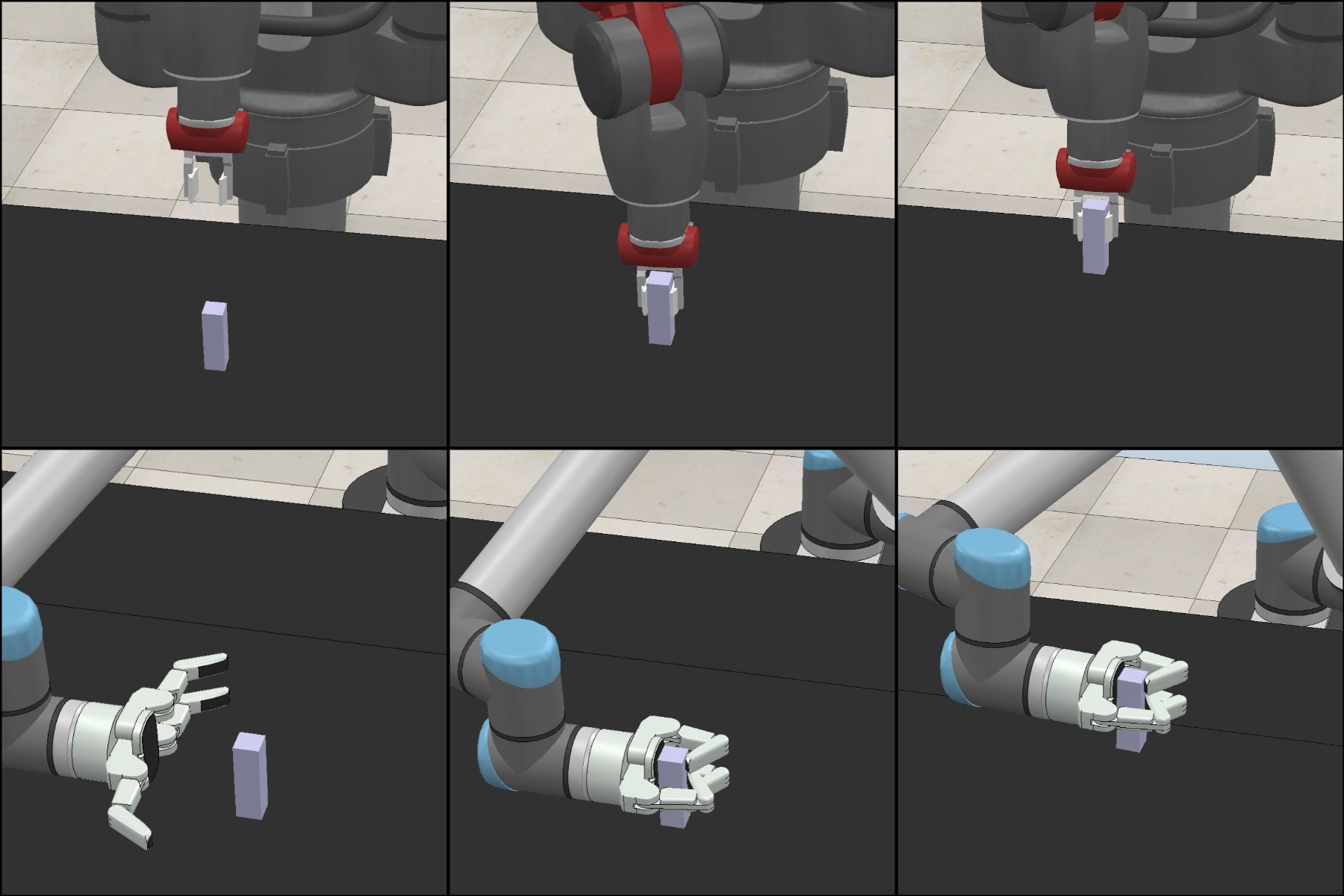}
    \caption{The grasp actions used in the experiment in Section \ref{section:grasp} are illustrated.}
    \label{figure:basic-grasp-setup}
\end{figure}

\begin{table*}[t]
    \centering
\begin{minipage}[t]{0.48\linewidth}\centering
\caption{The results of the experiment in Section \ref{section:grasp}. Inputs that correspond to graspable combinations were given to the system if possible. The errors are calculated using the trajectories and images that correspond to grasp.}
    \label{tab:graspable}
\scalebox{0.65}{
    \begin{tabular}{|c|c|c|c|c|c|c|c|c|c|c|c|} \hline 
         \makecell{input \\ configuration}&  \makecell{object+\\effect+\\UR10}&  \makecell{object+\\effect+\\baxter}&  \makecell{object+\\effect}&  \makecell{object+\\UR10}&  \makecell{object+\\baxter}&  \makecell{effect+\\UR10}&  \makecell{effect+\\baxter}&  object& effect&  UR10&  baxter\\ \hline 
         \makecell{image\\ error \\ (pixels)}&  0.0754&  0.0849&  0.0755&  0.0817&   0.0737&  0.195&  0.226&  0.0732&  0.199&  0.229&  0.231\\ \hline 
         \makecell{effect\\ error \\ (meters)}&  5.94e-4&   6.19e-4&  8.12e-4&  8.69e-4&  9.46e-4&  6.14e-4&   0.0906&  1.35e-3&  0.0912&   6.29e-4&    0.0898\\ \hline 
         \makecell{UR10\\ error \\ (radians)}&   0.0184&  0.0160&   0.0228&  0.0168&   0.0219&  0.0163&   0.0173&  0.0278&  0.0159&  0.0244&  0.0195\\ \hline 
         \makecell{baxter\\ error  \\ (radians)}&  0.0145&  0.0134&  0.0195&   0.0179&    0.0202&   0.0130&  0.0139&   0.0262&  0.0132&   0.0136&   0.0168\\ \hline
    \end{tabular}}
\end{minipage}\hfill%
\begin{minipage}[t]{0.48\linewidth}\centering
    \caption{The results of the experiment in Section \ref{section:grasp}. Inputs that correspond to non-graspable combinations were given to the system if possible. The errors are calculated using the trajectories and images that correspond to non-grasp.}
    \label{tab:non-graspable}
\scalebox{0.65}{ 
\begin{tabular}{|c|c|c|c|c|c|c|c|} \hline 
         \makecell{input \\ configuration} &  \makecell{object+\\effect+\\baxter}&  \makecell{object+\\effect}&  \makecell{object+\\baxter}&  \makecell{effect+\\baxter}&  object&  effect&   baxter\\ \hline 
         \makecell{image \\error \\ (pixels)}&  0.102&  0.102&   0.1&  0.103&  0.0991&   0.132&  0.101\\ \hline 
         \makecell{effect \\error \\ (meters)}&  4.14e-4&  4.36e-4&  8.43e-4&  3.84e-4&   7.2e-4&  8.24e-4&  9.94e-4\\ \hline 
         \makecell{UR10 \\error \\ (radians)}&  0.0164&   0.0179&  0.0181&  0.0173&   0.0187&   0.0159&  0.0195\\ \hline 
         \makecell{baxter\\ error \\ (radians)}&  0.0129&  0.0149&   0.0168&  0.0139&  0.0186&  0.0132&  0.0168\\ \hline
    \end{tabular}}
\end{minipage}
\end{table*}

The task is defined as grasping and lifting an object on the table. Two simulated robots (a UR-10 arm with a Barrett Hand and a Baxter robot with a Baxter gripper) are employed. The actions used by the robots are depicted in Figure \ref{figure:basic-grasp-setup}. Similar to the previous experiment, 10 objects are used, 5 can only be grasped by UR-10, while the remaining 5 can be grasped by both robots. 
Since the opening of the Baxter gripper is small, it can not grasp large objects. Out of these 10 objects, 8 are used for training, and 2 are used for testing, where the testing objects are not at the graspability boundary. \hakan{Since the elevation of the object center can be used to determine whether the object is successfully grasped and lifted, we defined the effect as the change in the position of the object.}


\hakan{Similar to the previous experiment, our system was able to generalize to different-sized objects that were not in the training set, and it was able to learn the affordances of different-sized objects using shared representations of the object equivalences. On the other hand,  while in the first experiment, one element of the affordance tuple was sufficient to infer the other elements, it is not always the case in this experiment. The generation error values using different input configurations can be seen in Tables \ref{tab:graspable} and \ref{tab:non-graspable} where Table \ref{tab:graspable} shows the results when the input indicates a graspable scenario and Table \ref{tab:non-graspable} shows the results when the input indicates a non-graspable scenario. For instance, when only the action of the Baxter or an object not graspable by either of the robots is given, the system can not generate a valid effect trajectory since the input does not contain enough information about the graspability. Therefore, the equivalence relation can be established between the agents only on the objects that are graspable by both agents and can be \erhan{denoted} with
$(grasped, \biggl\{   \begin{array}{l}
            UR-10\\
            Baxter
        \end{array}    ,(<smaller
Objects>, grasp  \biggl\} )$.}



\subsection{Cross-Embodied Transfer and Generalization of Rollability  on Novel Objects}\label{section:push}

\begin{figure}[b]
\centerline{\includegraphics[width=0.6\linewidth]{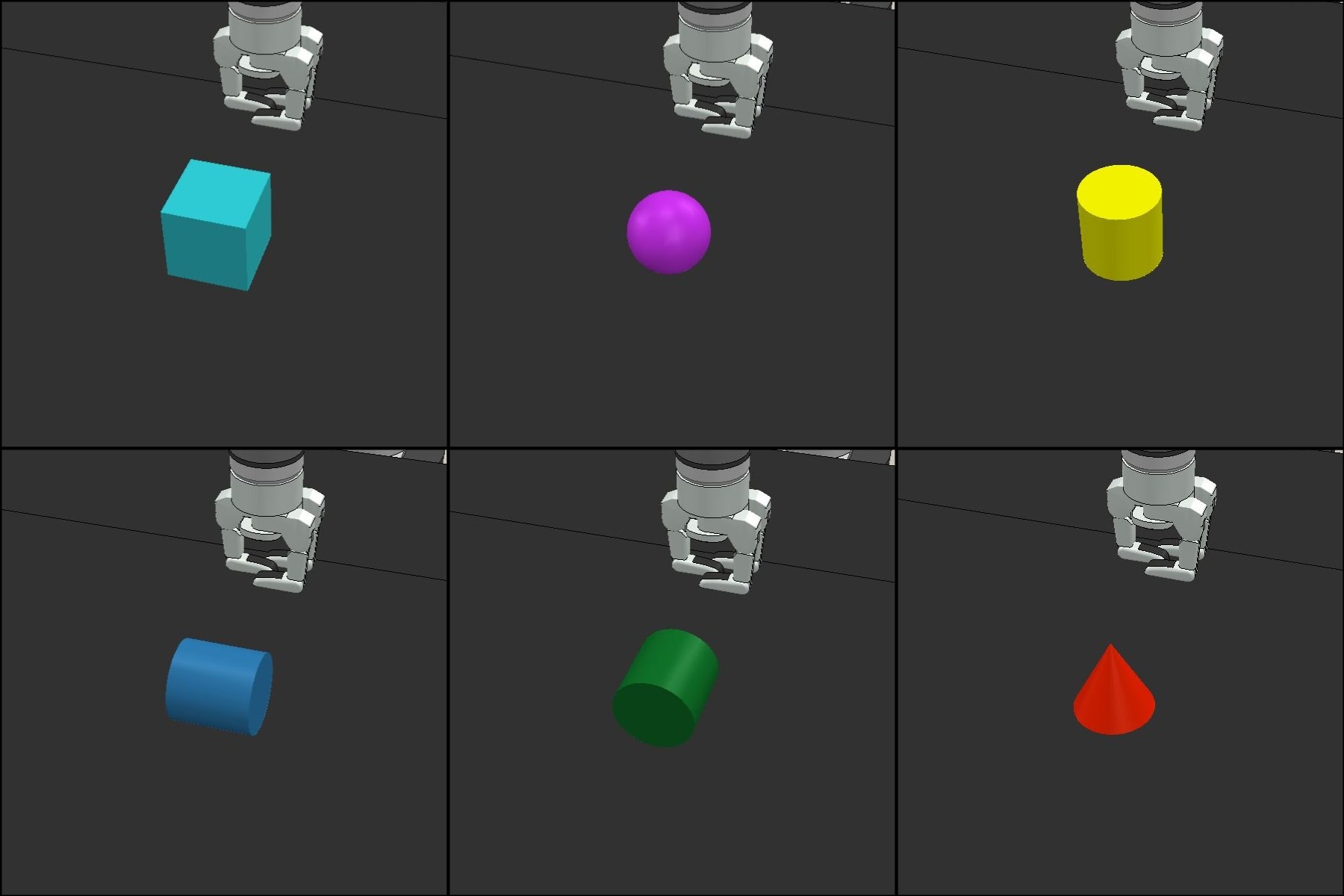}}
\caption{
The rollable and non-rollable objects used in the experiment in Section \ref{section:push}. While other objects have a fixed rollability (cuboid, sphere, and cone), a cylinder's rollability depends on its orientation. 
}
\label{figure:push-objects}
\end{figure}

\begin{figure}[b]
    \centering
    \includegraphics[width=1\linewidth]{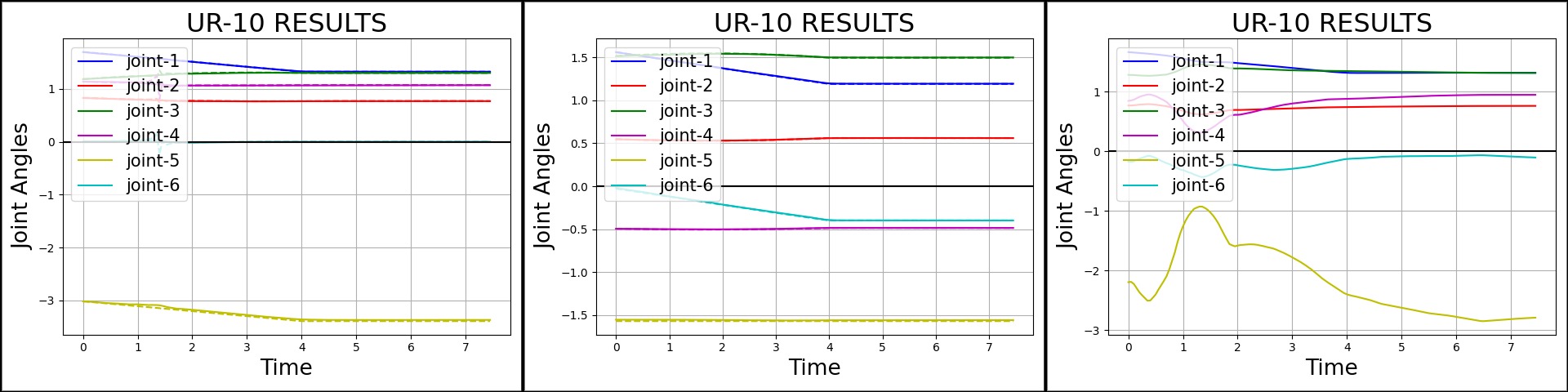}
        \caption{Generated action trajectories of UR-10 robot in the experiment in Section \ref{section:push}. The plot on the left and in the middle shows the generated actions when information about the UR-10 action is given.}
    \label{figure:push-results}
\end{figure}

\begin{figure*}[]
\centerline{\includegraphics[width=1\linewidth]{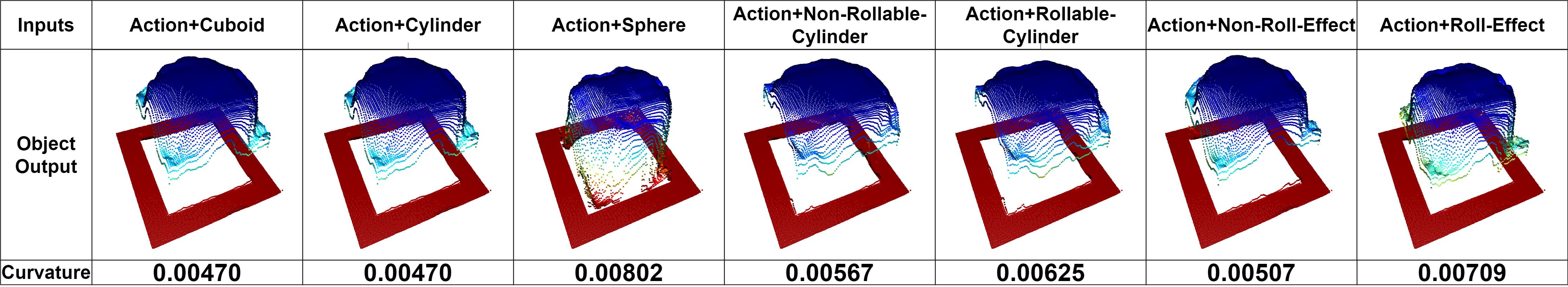}}
\caption{
The object depth images and average curvatures generated by different input conditions are shown as point clouds. A higher curvature generally indicates a more curved object.
}
\label{figure:all-point-clouds}
\end{figure*}

In this experiment, we aim to investigate how object rollability is learned by our model using two robotic agents. To this end,  simulated KUKA LBR4+ and UR-10 robots with two types of grippers (Robotiq85 and Barrett Hand) interact with objects with different rollability characteristics. The task for the robots is defined as pushing a given object in three possible directions (to the left, to the right, and straight). Two action types are used for UR-10 (one with its fingers while the gripper is closed and the other with the palm of the gripper).
Similar to the previous experiment, the effect is defined as the change in object position. The robots execute their push actions and wait for a fixed period to observe whether the object rolls.
The objects used in this experiment include a cuboid, an upright cylinder, a sphere, a cone, and cylinders on their sides with different orientations \hakan{and they can be seen in Figure \ref{figure:push-objects}}. While the rollability of some of them does not depend on the action applied to them (cuboid, upright cylinder, cone, and sphere), the rollability of the others (cylinders standing on their side) depends on the action's direction. All objects except the cone are used during training, and the cone is used to check whether our system can transfer affordances from one agent to another.

After training the model with all objects except the cone,
\erhan{the average curvatures of the generated objects are computed based on their point cloud representations to reason about the captured affordances.  }
The results showed that when an input configuration corresponding to rollable combinations is given to the system, the generated objects have a higher curvature than objects generated using non-rollable input combinations.
\hakan{The generated point clouds and their curvatures can be seen in Figure \ref{figure:all-point-clouds}. As seen, when the given input configuration indicates that the object is rollable, the curvature of the point cloud of the generated depth image is higher.}
The results also showed effect and action reconstruction accuracies similar to those of other experiments. However, since UR-10 has two different actions with the same outcome in this experiment when no information regarding its action is given, the system cannot generate a valid action trajectory for UR-10. One such trajectory can be seen on the right of Figure \ref{figure:push-results}. It can be seen that the generated trajectory \erhan{shows a different pattern compared to valid trajectories} (the ones on the left and at the center). \erhan{This case can be detected by the model as for such cases, the model outputs significantly higher variances at the UR-10 action channel, which can be used to prevent an invalid action execution.}

Additionally, we analyzed the system's ability to transfer a learned affordance from one robot to the other. For this, after training the model with a number of objects, we trained the model with the remaining objects with only one action of a single robot. After this additional training, when a new object image and the action of the robot the new object is not trained for is given to the system, it can predict the correct effect trajectory for that object. Furthermore, we checked whether the system generalized its rollability affordance to other push actions that were not used in the new demonstration. 
The generalization results can be seen in Table \ref{tab:rollability-phase-2}. The results showed that when the rollability affordance of the newly demonstrated object is present in the initial training objects, the affordance of the new object can be transferred to the other robots. However, if the new object exhibits a behavior that is unknown to the system, the system cannot transfer it (columns 4 and 7). On the other hand, generalization to other action directions only happens when no objects with different behaviors based on the direction of the action are present in the initial training objects since, if there are, the model learns that not all objects behave the same in all directions (column 2).

\begin{table}[t]
    \centering
    \scalebox{0.75}{
\begin{tabular}{|c|c|c|c|c|c|c|} \hline 
         initial training objects& \makecell{ all objects \\ except cone}&  \makecell{cuboid \\ cylinder} &  \makecell{cuboid \\ cylinder}&  \makecell{cuboid \\ cylinder \\ sphere}&  \makecell{cuboid \\ sphere}& sphere\\ \hline 
         new demonstration&  cone&  cone&  sphere&  cone&  cylinder& cone\\ \hline 
         \makecell{robot to robot \\ affordance   transfer}&  \checkmark &  \checkmark&  X&  \checkmark&  \checkmark& X\\ \hline 
         \makecell{action \\ direction generalization}&  X&  \checkmark&  X&  \checkmark&  \checkmark& X\\ \hline
    \end{tabular}}
    \caption{Affordance Transfer and Generalization Results}
    \label{tab:rollability-phase-2}
\end{table}

\begin{figure}[b]
    \centering
    \includegraphics[width=0.5\linewidth]{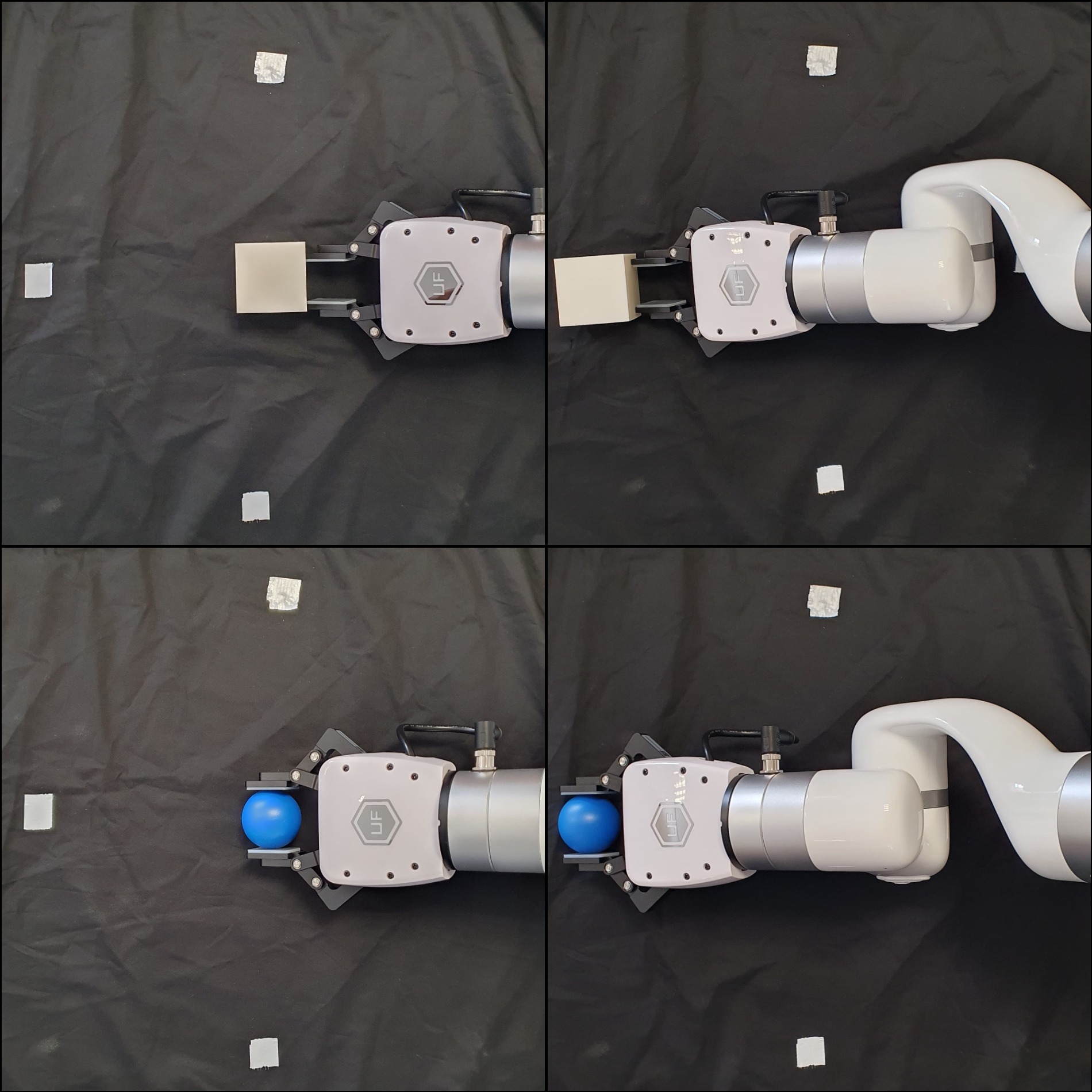}
    \caption{
The \emph{move}  actions of the robots (Section \ref{section:real-robot}).}
    \label{figure:real-robot}
\end{figure}


\subsection{Effect-Based Direct Imitation with Real Robot} \label{section:real-robot}
Since our system can generate action trajectories using effect trajectories and object information, it can be used for direct imitation. We designed an experiment using a real xArm 7 robot equipped with an xArm gripper to test this ability. The task is defined as bringing a given object to a desired location. Both rollable and non-rollable objects are used, and different actions are defined for them (see Figure~\ref{figure:real-robot}). The locations of the objects are tracked using a depth camera. During testing, the experimenter pushes one of the objects towards a particular location. The object's final position is given to the system along with the depth image of the object, and action generation is requested. The robot executes the generated action trajectory after placing the object back in its initial position; our model is equipped with the robot with the ability to perform effect-based imitation. See the included video for the execution performance of our model (\url{https://youtu.be/U6swpEDsaC0}). 
\section{Conclusion, Limitations, and Future Work}
\label{sec:conclusion}
We developed a model that can learn affordance representations that can be used to generate continuous action and effect trajectories and object images that constitute those affordances. Our system could learn common representations for equivalent representations for the affordances we used in the experiments. 
 We analyzed the formation of these representations using a latent space analysis. We also showed that affordances could be used to couple different robots' actions, allowing cross-embodiment transfer. We also demonstrated our model's generalization capabilities in transferring affordance information between agents. Moreover, in real-world experiments, we showed that our model can be used for direct imitation. 
 
 Some limitations of our model include 
 \erhan{the inefficiency in introducing more robots to the system after  training with the initial set of robots. Currently, the only way to achieve this is to retrain the whole system with the new robot data added. As for the cross-embodiment transfer, the model relies on an initial data set collected from agents when they perform the same task.  This should be addressed in future iterations of the model. Another assumption that should be relaxed in future studies is that all the actions are represented with the same length of time series data, which may limit the generality of the actions that can be considered.}

 \erhan{Finally, a fruitful venue for improvement is to address how the model behaves when it is given an ambiguous input in the sense that there are multiple compatible outputs. In these cases, the model is able to detect the issue and report it. However, in some applications, it may be desirable that the system picks one of the alternatives and outputs that as the answer. Last but not least, the use of robots with significantly different morphologies with possible multiple effect channels should be investigated as  future work.}

\section*{ACKNOWLEDGMENT}

This research has been funded by the JST Moonshot R\&D, Japan (JPMJMS2292), by the JST CREST, Japan (JPMJCR21P4), by the JSPS KAKENHI, Japan (21H05053), by the World Premier International Research Center Initiative (WPI), MEXT, Japan,  by JSPS KAKENHI Grant Number JP90542217, the project JPNP16007 commissioned by the New Energy and Industrial Technology Development Organization (NEDO), and the Scientific and Technological Research Council of Turkey (TUBITAK, 118E923). This work was also supported by the INVERSE Project under Grant 101136067 funded by the European Union.

  \bibliographystyle{IEEEtran}
 \bibliography{references}

\end{document}